\documentclass[letterpaper]{article} % DO NOT CHANGE THIS
\usepackage{aaai24}  % DO NOT CHANGE THIS
\usepackage{times}  % DO NOT CHANGE THIS
\usepackage{helvet}  % DO NOT CHANGE THIS
\usepackage{courier}  % DO NOT CHANGE THIS
\usepackage[hyphens]{url}  % DO NOT CHANGE THIS
\usepackage{graphicx} % DO NOT CHANGE THIS
\usepackage{natbib}  % DO NOT CHANGE THIS AND DO NOT ADD ANY OPTIONS TO IT
\usepackage{caption} % DO NOT CHANGE THIS AND DO NOT ADD ANY OPTIONS TO IT
\DeclareCaptionStyle{ruled}{labelfont=normalfont,labelsep=colon,strut=off} % DO NOT CHANGE THIS
\usepackage{float}
\usepackage{tabularx}
\usepackage{multirow}
\usepackage{amsmath}
\usepackage{bm}
\usepackage{arydshln}
\usepackage{amsfonts}
\usepackage{pgfplots}
\pgfplotsset{compat=1.7}
\usepackage{makecell}
\usepackage{subfig}                 % 子图包，不要与{subfigure}混用，{subfig}较新
\usepackage{overpic}                % 叭啦叭啦，与图片排版相关
\usepackage{pifont}
\usepackage{colortbl}
\usepackage{algorithmic}
\usepackage[ruled,vlined]{algorithm2e}
\usepackage{booktabs}       % professional-quality tables

\newcommand{\method}{Proto-semi}

\title{Rethinking Noisy Label Learning in Real-world Annotation Scenarios from the Noise-type Perspective}
\author{
  Renyu~Zhu\textsuperscript{1}, Haoyu~Liu\textsuperscript{1}, Runze~Wu\textsuperscript{1}, Minmin~Lin\textsuperscript{1},
  Tangjie~Lv\textsuperscript{1}, Changjie~Fan\textsuperscript{1}, Haobo~Wang\textsuperscript{2}\\
  \textsuperscript{1}NetEase~Fuxi~AI~Lab,
  Hangzhou, Zhejiang, China \\
  \texttt{\{zhurenyu, liuhaoyu03, wurunze1, linminmin01, hzlvtangjie, fanchangjie\}@corp.netease.com}\\
  \textsuperscript{2}Zhejiang Univeristy, Hangzhou, Zhejiang, China \\
  wanghaobo@zju.edu.cn \\
}
\usepackage{bibentry}
\begin{document}

\maketitle
\begin{abstract}
In this paper, we investigate the problem of learning with noisy labels in real-world annotation scenarios, where noise can be categorized into two types: factual noise and ambiguous noise. To better distinguish these noise types and utilize their semantics, we propose a novel sample selection-based approach for noisy label learning, called \method. \method\ initially divides all samples into the confident and unconfident datasets via warm-up. By leveraging the confident dataset, prototype vectors are constructed to capture class characteristics. Subsequently, the distances between the unconfident samples and the prototype vectors are calculated to facilitate noise classification. Based on these distances, the labels are either corrected or retained, resulting in the refinement of the confident and unconfident datasets. Finally, we introduce a semi-supervised learning method to enhance training. Empirical evaluations on a real-world annotated dataset substantiate the robustness of \method\ in handling the problem of learning from noisy labels. Meanwhile, the prototype-based repartitioning strategy is shown to be effective in mitigating the adverse impact of label noise. Our code and data are available at \url{https://anonymous.4open.science/r/ProtoSemi-C61F}.

\end{abstract}

\section{Introduction}

The rapid development of deep learning technologies such as ChatGPT~\cite{DBLP:conf/nips/Ouyang0JAWMZASR22} is inseparable from the support of high-quality labeled data. 
Unfortunately, 
creating such data is costly as it requires expert annotation.
To address this issue, 
researchers have introduced low-cost data collection methods such as crowdsourcing~\cite{DBLP:journals/cacm/DoanRH11}. 
However, 
the quality of the collected data is hard to fully guarantee, 
resulting in the production of noisy labels.

The occurrence of noisy labels may cause the model to suffer from the memorization phenomenon~\cite{DBLP:conf/icml/ArpitJBKBKMFCBL17, DBLP:conf/nips/BaiYHYLMNL21}, 
which leads to overfitting and performance decrease. 
As a result, 
numerous works~\cite{DBLP:journals/corr/abs-2303-10802, DBLP:journals/corr/abs-2007-08199} have been devoted to learning from noisy labels, 
such as robust structures~\cite{DBLP:conf/iclr/GoldbergerB17, DBLP:journals/tip/YaoWTZSZZ19},
robust regularization~\cite{DBLP:conf/icdm/JindalNC16, DBLP:conf/iclr/ZhangCDL18}, 
robust loss functions~\cite{DBLP:conf/icml/ArazoOAOM19, DBLP:conf/cvpr/PatriniRMNQ17},
and sample selection-based methods~\cite{DBLP:conf/nips/BaiYHYLMNL21, DBLP:conf/iclr/LiSH20}. 
% We provide a detailed overview of these methods in Section~\ref{sec:related_works_noisy}.
Among them, 
the sample selection-based methods first identify clean samples with true labels from all samples,
and introduce semi-supervised learning~\cite{DBLP:conf/nips/BerthelotCGPOR19} to enhance training,
which achieve state-of-the-art performance in noisy label learning.

% \cite{DBLP:journals/corr/abs-2007-08199}

\begin{figure}[t]
	\centering
	\subfloat[Factual noise]{\includegraphics[width=.99\columnwidth]{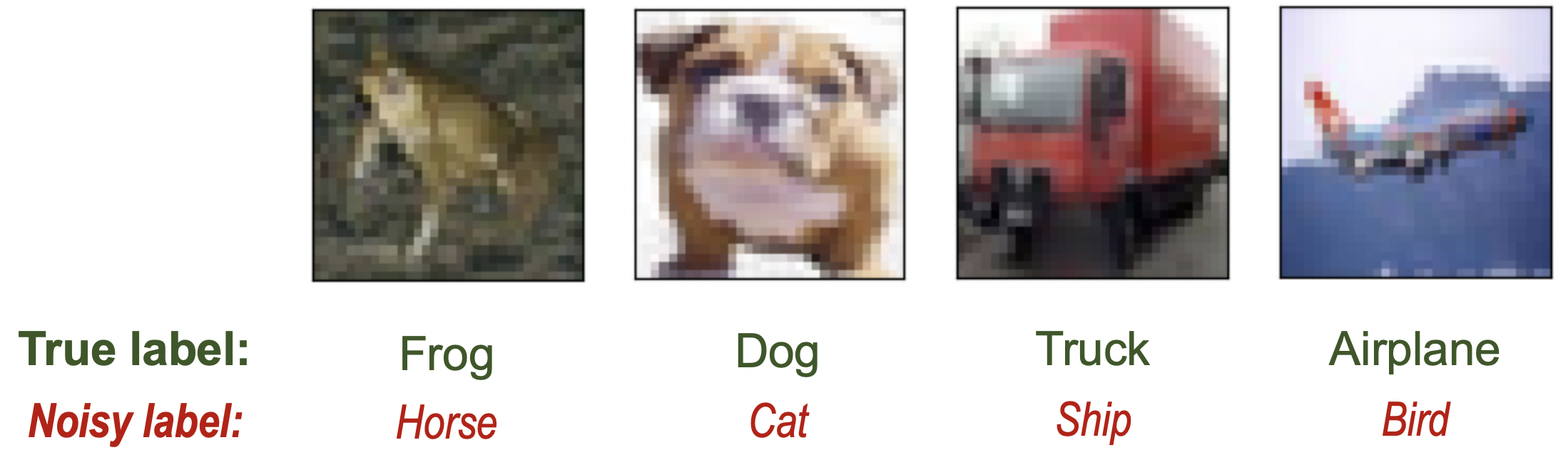}}\\
	\subfloat[ambiguous noise]{\includegraphics[width=.99\columnwidth]{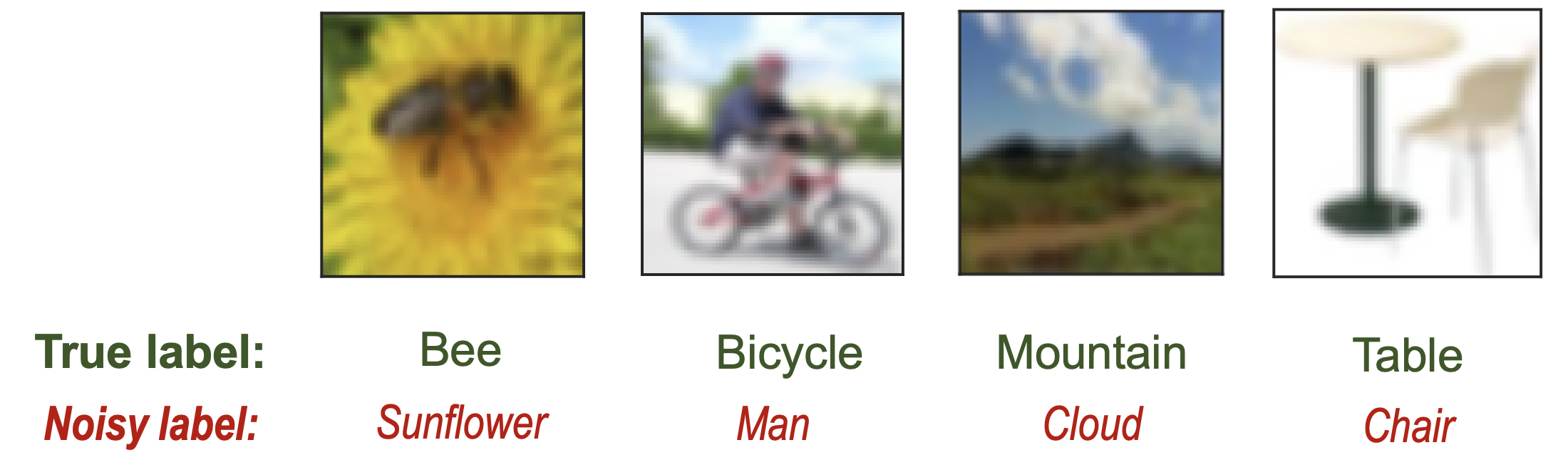}}
	\caption{Examples of the two noise types in real-world annotation scenarios.}
 \label{fig:noise_type}
\end{figure}

Meanwhile, 
previous works~\cite{DBLP:journals/corr/abs-2007-08199, DBLP:conf/iclr/WeiZ0L0022} have indicated that noise generated during annotation in real-world scenarios can be divided into two types:  
1) \textbf{Factual noise}: 
the annotator mistakenly labels the sample due to carelessness or lack of knowledge during the annotation process; 
2) \textbf{ambiguous noise}: 
the annotator labels the sample incorrectly due to its ambiguity, such as multiple labels.
The examples of the two noise types are shown in Figure~\ref{fig:noise_type}.
% these labels reflect information about the sample in another dimension.
However, 
previous sample selection-based methods~\cite{DBLP:conf/nips/BaiYHYLMNL21, DBLP:conf/iclr/LiSH20} neglect the noise types and simply remove the labels of all noisy samples.
It inspires us to rethink noise label learning in real-world annotation scenarios: \textit{How to distinguish the noise types and utilize their semantics?}

% \begin{figure}
%      \centering
%      \begin{subfigure}[b]{0.48\textwidth}
%         \centering
%         \includegraphics[height=2.5cm]{figs/factual.png}
%         \caption{Factual noise}
%         \label{fig:type_factual}
%      \end{subfigure}
%      \hfill
%      \hfill
%      \begin{subfigure}[b]{0.48\textwidth}
%         \centering
%         \includegraphics[height=2.5cm]{figs/ambiguous.png}
%         \caption{ambiguous noise}
%          \label{fig:type_ambiguous}
%      \end{subfigure}
%         \caption{Examples of the two noise types in real-world annotation scenarios.}
%         \label{fig:noise_type}
% \end{figure}

To answer the question,
in this paper, 
we propose a novel sample selection-based method,
called \method,
for noisy label learning in real-world annotation scenarios.
Similar to previous sample selection-based methods~\cite{DBLP:journals/corr/abs-2303-10802, DBLP:journals/corr/abs-2007-08199, DBLP:conf/iclr/WeiZ0L0022}, 
We first perform a warm-up pretraining to divide the training samples into confident and unconfident datasets.
Then we create a prototype vector for each class based on the corresponding samples in the confident dataset, 
and calculate the distance between each sample in the unconfident dataset and these prototypes. 
After that, the noise types of samples in the unconfident dataset are identified based on the calculated distance:
1) For samples with a small distance, we consider them to be factual noise and perform label correction.
2) For samples with a distance that is neither too large nor too small, 
we categorize it as an ambiguous error and correct or retain the label based on the probability according to the distance value. 
The corrected or retained data, along with their labels, are moved to the confident dataset.
% the distance between each sample vector and the prototype vector for each class is computed, 
% while two distance thresholds are set, one low and one high. 
% If the distance is less than the low threshold but the sample has a label different from the prototype vector, 
% it is marked as a factual error. 
% Conversely, if the distance is between the low and high thresholds but the sample has a label different from the prototype vector, 
% it is marked as an ambiguous error. For factual errors, we directly correct the labels.
% While for ambiguous errors, 
% we apply a probability based on the distance to either correct or retain the label.
Finally, we introduce the semi-supervised learning method~\cite{DBLP:conf/nips/BerthelotCGPOR19} to train the model based on the refined confident and unconfident datasets. In summary, our contributions are followed as:
\begin{itemize}
    \item We present a new noise-type perspective on noisy label learning in real-world scenarios.
    \item We propose a novel sample selection-based method called \method\ for noisy label learning. \method\ introduces prototypes to repartition confident and unconfident datasets based on the types of noise.
    \item Experiments on the CIFAR-N~\cite{DBLP:conf/iclr/WeiZ0L0022} dataset demonstrate the effectiveness of \method\ for noisy label learning in real-world scenarios. Additionally, we use prototype visualization and detailed statistical analysis to confirm that \method\ can effectively identify the type of noise and correct the labels for factual noise.
\end{itemize}

% ~\cite{DBLP:conf/nips/HanYNZTZS18, DBLP:journals/tip/YaoWTZSZZ19}
% ~\cite{DBLP:conf/iccv/ChenG15, DBLP:conf/iclr/GoldbergerB17}

\section{Related Works}

\subsection{Learning From Noisy Labels}\label{sec:related_works_noisy}
The existing literature on noisy label learning can be broadly classified into four categories. The first category focuses on designing more robust architectures, such as dedicated architectures~\cite{ DBLP:conf/cvpr/XiaoXYHW15, DBLP:journals/tip/YaoWTZSZZ19} and the addition of noise adaptation layers~\cite{DBLP:conf/iccv/ChenG15, DBLP:conf/iclr/GoldbergerB17} at the top of the softmax layer to model the noise transition matrix. The second category centers on robust regularization, including explicit~\cite{DBLP:conf/icml/HendrycksLM19, DBLP:conf/cvpr/TannoSSAS19} and implicit regularization~\cite{DBLP:journals/corr/GoodfellowSS14, DBLP:conf/icml/LukasikBMK20}, both of which aim to prevent overfitting in noisy data. The third category focuses on developing more robust loss functions, such as loss regularization~\cite{DBLP:conf/iclr/LyuT20}, loss correction~\cite{DBLP:conf/cvpr/PatriniRMNQ17}, and loss reweighing~\cite{DBLP:journals/pami/LiuT16}. Finally, the last category is sample selection-based~\cite{DBLP:conf/nips/BaiYHYLMNL21, DBLP:conf/iclr/LiSH20}, which involves selecting true labeled examples from noisy training datasets and has been shown to be the most effective method among all categories.

Sample selection-based methods can be further classified into two categories: loss-based sampling~\cite{DBLP:conf/wacv/SachdevaCBRC21, DBLP:conf/wacv/ZheltonozhskiiB22} and feature-based sampling~\cite{DBLP:conf/nips/KimKCCY21, DBLP:conf/nips/WuZ0M020}. Loss-based sample selection methods are based on the small loss hypothesis, which assumes that noisy data tends to incur a high loss due to the model's inability to accurately classify such data. Feature-based sampling methods distinguish clean and noisy samples by leveraging features. However, both of these categories simply categorize all predicted noisy samples as unlabeled, which does not fully exploit the training data. Therefore, in this paper, we propose a novel noise-type-based sample selection method to enhance the utilization of annotation information in training data.

\subsection{Prototype Neural Networks}
Prototype-based metric learning methods~\cite{DBLP:conf/nips/SnellSZ17} have emerged as a promising technique in numerous research domains, including few-shot learning~\cite{DBLP:journals/csur/WangYKN20} and meta-learning~\cite{DBLP:journals/air/VilaltaD02}. ProtoNet~\cite{DBLP:conf/nips/SnellSZ17}, which introduces prototypes into deep learning, is a seminal work in this area. Subsequent studies~\cite{DBLP:conf/cvpr/LiJSSKK21, DBLP:conf/iclr/0001ZXH21} have focused on improving prototype estimation to enhance classification performance. The success of prototype-based models suggests that prototypes can serve as representative embeddings of instances from the same classes, thereby inspiring us to explore the use of prototype networks in the context of noisy label learning.

\section{Problem Definition}

In practical problems, it is often difficult to obtain samples from the true distribution $D$ of a pair of random variables $(\bm{X}, \bm{Y})\in\mathcal{X}\times \{1,…K\}$ due to random label corruption. Here, $\bm{X}$ is the sample variable, $\bm{Y}$ is the label variable, $\mathcal{X}$ represents the feature space, and $K$ is the number of classes. Instead, in the context of noisy label learning, we can only access a set of samples $\{{({\bm{x}_{i}, \tilde{y}_{i}})_{i = 1}^{n}}\}$ that are independently drawn from the distribution $\tilde{D}$ of noisy samples $(\bm{X}, \bm{\tilde{Y}})$, where $\bm{\tilde{Y}}$ represents the noisy label variable. The objective is to learn a robust classifier based on noisy samples that can accurately classify test samples.

\section{Method}
In this section, we introduce our proposed method \method\ for learning with noisy labels in real-world annotation scenarios. \method\ is mainly composed of two steps: sample selection and semi-supervised learning. Sample selection is used to divide the whole dataset into the confident and unconfident datasets, in which we propose a prototype-based repartitioning strategy to identify the types of noise and correct their labels. And then semi-supervised learning is introduced to enhance the training based on the repartitioned confident and unconfident datasets. Details of the two steps are described in the following.
% Firstly, the process of the true labeled examples selection in the \method\ is presented. Then we give the complete training workflow of \method.

\subsection{Sample Selection}
% First, 
% As demonstrated by the previous works~\cite{}, 
% deep neural networks will first 

Similar to previous label-noise learning methods based on sample selection~\cite{DBLP:journals/corr/abs-2303-10802, DBLP:journals/corr/abs-2007-08199, DBLP:conf/iclr/WeiZ0L0022}, we begin by dividing the original noisy dataset $\{({\bm{x}_{i}, \tilde{y}_{i}})_{i = 1}^{n}\}$ into two distinct subsets: a labeled confident dataset $\mathcal{D}_{l}$ and an unlabeled unconfident dataset $\mathcal{D}_{u}$. In this work, we accomplish this through a warm-up phase, during which we optimize a training function as follows:
\begin{align}
& \min_{\Theta}\frac{1}{n}\sum_{i = 1}^{n}\mathcal{L}(f(\bm{x}_{i};\Theta), {\tilde{y}_{i}}),
\label{equ:warm_up}
\end{align}
where $f(\cdot; \Theta)$ denotes the classifier, such as deep neural networks, $\Theta$ is the parameters of classifier, and $\mathcal{L(\cdot)}$ is the cross-entropy loss function. To avoid the memorization phenomenon~\cite{DBLP:conf/icml/ArpitJBKBKMFCBL17}, we only perform a small number of epochs in the warm-up phase. Upon completing the warm-up phase, we proceed to identify samples across all examples whose classifier soft outputs are consistent with their current labels. The identified samples are put into the confident dataset $\mathcal{D}_{l}$ and their labels are retained. In contrast, samples whose soft outputs and current labels are inconsistent are dropped into unconfident dataset $\mathcal{D}_{u}$ and their labels are discarded. In short, the split process can be summarized by the following two formulas:
\begin{align}
    \mathcal{D}_{l} = & \{({\bm{x}_{i}, \tilde{y}_{i}})|\tilde{y}_{i} = \hat{y}_i, i = 1, \cdots, n\},\label{equ:conf}\\
    \mathcal{D}_{u} = & \{\bm{x}_{i}| \tilde{y}_{i} \neq \hat{y}_i, i = 1, \cdots, n\}\label{equ:unconf},
\end{align}
where $\hat{y}_i$ represents the soft output of classifier $f(\cdot; \Theta)$ on sample $i$. As pointed out by previous works~\cite{DBLP:conf/nips/BaiYHYLMNL21}, during the warm-up phase, the model gradually converges to clean samples that make up the majority of all samples\footnote{In the CIFAR-N dataset studied in this paper, the highest level of noise is approximately 40\%.}. Therefore, based on the warm-up dividing, the majority of clean samples and noisy samples are assigned to the confident dataset $\mathcal{D}_{l}$ and unconfident dataset $\mathcal{D}_{u}$, respectively. However, as aforementioned in the Introduction, noise in the human annotation scenario can be mainly divided into factual noise and ambiguous noise. To distinguish these noises and utilize the information effectively, we propose a prototype-based repartitioning strategy.

First, we construct an $M$-dimensional representation $\bm{c}_k \in \mathcal{R}^{M}$, or \textit{prototype}, of each class through an embedding function. In this paper, we use the classifier after warm-up and remove its last linear layer as the embedding function, tagged as $f^{'}(\cdot;\Theta)$. And each prototype is the mean vector of all sample embeddings belonging to its class:
\begin{align}
    \bm{c}_k & = \frac{1}{|D_l^k|} \sum_{(\bm{x}_i, \hat{y}_i) \in D_l^k} f^{'}(\bm{x}_i; \Theta), \label{equ:proto}
\end{align}
where $D_l^k$ denotes the all samples that belong to the class $k$ in the confident dataset $D_l$, and $|D_l^k|$ is the number of samples in $D_l^k$. Similarly, we can obtain the prototype vectors of classes in addition to the class $k$ based on Eq.~\eqref{equ:proto}. And we aggregate all the prototype vectors of each class together to obtain the prototype matrix $\mathbf{C}$.

After that, for each sample $u$ in the unconfident dataset $D_u$, we calculate the distance between the sample and all prototypes:
\begin{align}
    \bm{d} & = Distance(f^{'}(u; \Theta), \mathbf{C})\label{equ:dis},
\end{align}
where $\bm{d}$ denotes the distance vector and $Distance(\cdot, \cdot)$ represents the distance function. We use cosine similarity as the distance function in this paper.

\begin{figure}[h]
    \centering
    \includegraphics[width=0.48\textwidth]{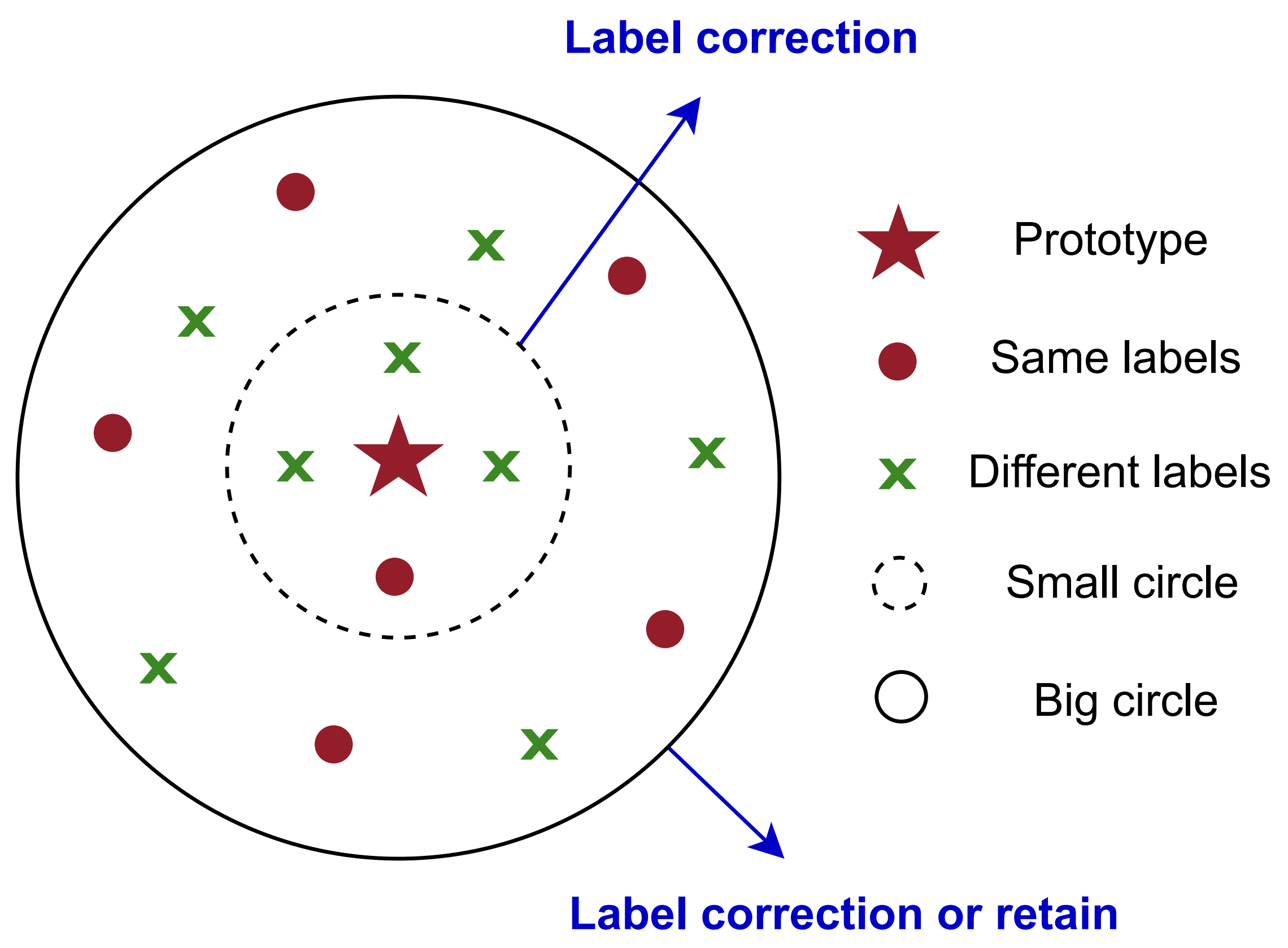}
    \caption{Noise type identifying in \method.}
    \label{fig:dis}
\end{figure}

% \begin{wrapfigure}[17]{r}{7.5cm}
% \centering
% \includegraphics[width=0.5\textwidth]{figs/main-new.png}
% \caption{Noise type identifying in \method.}
% \label{fig:dis}
% \end{wrapfigure}

Then we select the maximum distance $d_{max}$ from the distance vector $\bm{d}$ and assign the class with the highest value as the \textit{proto-label}~\footnote{Particularly, in this paper, we perform data augmentation twice for each sample, and only when the positions of maximum similarity in the two distance vectors are the same, we construct the \textit{proto-label} for the current sample.}. Note that each sample in the unconfident dataset also has its original label, referred to as the \textit{noisy-label}. Based on the prototype of the \textit{proto-label}, two circles are constructed as shown in the Figure~\ref{fig:dis}: 1) \textbf{Small circle}: the distance to the prototype is less than the lower threshold $\alpha$. 2) \textbf{Big circle}: the distance to the prototype is less than the upper threshold $\beta$. Based on the two circles, we are able to determine which samples should be moved from the unconfident dataset $D_u$ to the confident dataset $D_l$:
\begin{itemize}
    \item \textbf{The sample that fall within the small circle ($d_{max} < \alpha$)}: The sample will be assigned the \textit{proto-label}. If the \textit{noisy-label} is different from the \textit{proto-label}, this indicates the factual noise is identified and corrected.
    \item \textbf{The samples that lie between the small and big circles ( $\alpha 
    \leq d_{max} < \beta$)}: If the \textit{noisy-label} is equal to the \textit{proto-label}, the sample's original label, namely \textit{noisy-label}, will be retained; Otherwise, the sample will be corrected to the  \textit{proto-label} probabilistically based on the distance to the prototype, with the probability of correction increasing as the distance decreases. This randomness allows us to correct factual noises while also utilizing semantic information from ambiguous noises.
\end{itemize}

% Samples that fall within the small circle are assigned the same label as the prototype. In this case, factual errors are identified and corrected, while the correct labels are retained. For samples that lie between the small and big circles, the same labels are retained, but different labels are corrected or retained probabilistically based on the distance to the prototype. The probability of correction increases as the distance decreases, allowing us to correct factual errors while also utilizing semantic information from ambiguous errors.

Finally, all unconfident samples that fall within the big circle (including the small circle), after label correction or retention, will be moved to the confident dataset, creating new confident and unconfident datasets.

\subsection{Semi-supervised Learning}
After the new confident dataset $D_l$ and unconfident dataset $D_u$ are generated, inspired by the previous noisy label learning works~\cite{DBLP:journals/corr/abs-2303-10802, DBLP:journals/corr/abs-2007-08199, DBLP:conf/iclr/WeiZ0L0022}, we introduce the semi-supervised training method, MixMatch~\cite{DBLP:conf/nips/BerthelotCGPOR19}, to enhance the training in \method. 
The procedure of semi-supervised learning used in \method\ is as follows.

We begin by augmenting the data in both the refinement confident dataset $D_l$ and the unconfident dataset $D_u$, resulting in the new augmented datasets $D_{l}^{'}$ and $D_{u}^{'}$, respectively. 
Following the approach described in~\cite{DBLP:conf/nips/BaiYHYLMNL21}, we apply the augmentation technique with two different strategies to each dataset:
\begin{align}
  D_{l1}^{'} & = \text{\textit{Augment}}_{1}(D_l),  D_{l2}^{'} = \text{\textit{Augment}}_{2}(D_l), \label{equ:conf_aug}\\
  D_{u1}^{'} & = \text{\textit{Augment}}_{1}(D_u),  D_{u2}^{'} = \text{\textit{Augment}}_{2}(D_u),\label{equ:unconf_aug}
\end{align}
where \textit{Augment}$_{(\cdot)}$ denotes the augmentation technique. 

Particularly, we use the aggregated model soft outputs of $D_{u1}^{'}$ and $D_{u2}^{'}$ to guess the pseudo-labels for the augmented unconfident dataset. After that, Mixup~\cite{DBLP:conf/iclr/ZhangCDL18} is incorporated to combine the inputs and labels in two confident datasets ($D_{l1}^{'}$, $D_{l2}^{'}$) and unconfident datasets ($D_{u1}^{'}$, $D_{u2}^{'}$), respectively.
Finally, for the combined confident dataset, we use the \textit{cross-entropy} loss $\mathcal{L}_{l}$ to evaluate the difference between the outputs and true labels. While for the combined unconfident dataset, the \textit{L2} loss $\mathcal{L}_{u}$ is introduced to evaluate the discrepancy between the outputs and pseudo-labels. And the final training loss is the weighted sum of the two losses:
\begin{align}
  \mathcal{L}_{final} & = \mathcal{L}_{l} +  \lambda \mathcal{L}_{u}\label{equ:loss}, 
\end{align}
where $\lambda$ is the balance factor.

We summarize the overall procedure of \method\ in Algorithm~\ref{algorithm1}.

\begin{algorithm}[h]
{\bfseries Input}:  Neural network with trainable parameters $\Theta$, Noisy training dataset $\{(\bm{x}_{i}, \tilde{y}_{i})\}_{i = 1}^{n}$, Number of warm-up epochs $T_{w}$, proto split epochs $T_{p}$, and training epochs $T_{t}$ for refining with confident examples and unconfident examples, Lower threshold $\alpha$, Upper threshold $\beta$. 

\For{$i = 1, \dots, T_{w}$ }{
     {Optimize network parameter} $\Theta$ with Eq.~\eqref{equ:warm_up}; \\
}

\For{$i = 1, \dots, T_{c}$ }{

{Extract} confident example set $\mathcal{D}_{l}$ and unlabeled set $\mathcal{D}_{u}$ based on classifier $f(\cdot, \Theta)$ with Eq.~\eqref{equ:conf} and Eq.~\eqref{equ:unconf};\\
\If{$i < T_{p}$}{
    % {Introduce} proto split to obtain the new $\mathcal{D}_{l}$ and $\mathcal{D}_{u}$;\\
    {Contruct} prototype for each class based on $\mathcal{D}_{l}$ with Eq.~\eqref{equ:proto}; \\
    \For{$u \in \mathcal{D}_{u}$}{
        {Calculate} the distances between $u$ and all class prototypes with Eq.~\eqref{equ:dis}; \\
        {Select} the maximum value $d_{max}$ in all distances; \\
        \If{$d_{max} \leq \alpha$}{
            {Assign} the class corresponding to $d_{max}$ as the label of $u$; \\
            {Move} $u$ from $\mathcal{D}_{u}$ to $\mathcal{D}_{l}$; \\        
        }
        \If{$d_{max} > \alpha \And d_{max} \leq \beta$}{
            {Assign} the class corresponding to $d_{max}$ as the label of $u$ or retain the original label of $u$ with regard to the  value of $d_{max}$; \\
            % {Assign} the class corresponding to $cos_{max}$ as the label of $u$;
            {Move} $u$ from $\mathcal{D}_{u}$ to $\mathcal{D}_{l}$ ; \\
        }
        
    }

}
	 {Training} the classifier $f(\cdot, \Theta)$ with MixMatch loss on the new generated $\mathcal{D}_{l}$ and $\mathcal{D}_{u}$ with Eq.~\eqref{equ:loss};}\

{Evaluate} the obtained classifier $f(\cdot, \Theta)$. \\
\caption{The Overall Procedure of \method}
\label{algorithm1}
\end{algorithm}

% \begin{figure}[t]
%     \centering
%     \includegraphics[width=0.5\textwidth]{figs/main.png}
%     \caption{The error-type judging in \method.}
%     \label{fig:main}
% \end{figure}

\section{Experiments}

\begin{table*}[h]
  \caption{The classification results (\%) over the methods on the CIFAR-N dataset w.r.t. accuracy. 
  We highlight the best results in bold and the runner-up results with underline. }
  \label{tab:res_main}
  \centering
  \begin{tabular}{c|ccccc|c}
  \toprule
   \multirow{2}*{Methods} & \multicolumn{5}{c}{CIFAR-10N}  &  CIFAR-100N \\
   & aggre & rand1 & rand2 & rand3 & worst & noisy  \\
    \midrule
    Co-Teaching & 91.20$\pm$0.13 & 90.33$\pm$0.13 & 90.30$\pm$0.17 & 90.15$\pm$0.18 & 83.83$\pm$0.13 & 60.37$\pm$0.27 \\
    Peer-Loss & 90.75$\pm$0.25 & 89.06$\pm$0.11 & 88.76$\pm$0.19 & 88.57$\pm$0.09 & 82.53$\pm$0.52 & 57.59$\pm$0.61 \\
    Divide-Mix & 95.01$\pm$0.71 & 95.16$\pm$0.19 & 95.23$\pm$0.07 & 95.21$\pm$0.14 & 92.56$\pm$0.42 & $\mathbf{71.13\pm0.48}$ \\
    ELR & 92.38$\pm$0.64 & 91.46$\pm$0.38 & 91.61$\pm$0.16 & 91.41$\pm$0.44 & 83.58$\pm$0.13 & 58.94$\pm$0.92\\
    ELR+ & 94.83$\pm$0.10 & 94.43$\pm$0.41 & 94.20$\pm$0.24 & 94.34$\pm$0.22 & 91.09$\pm$1.60 & 66.72$\pm$0.07\\
    Positive-LS & 91.57$\pm$0.07 & 89.80$\pm$0.28 & 89.35$\pm$0.33 & 89.82$\pm$0.14 & 82.76$\pm$0.53 & 55.84$\pm$0.48\\ 
    PES-semi & 94.66$\pm$0.18 & 95.06$\pm$0.15 & 95.19$\pm$0.23 & 95.22$\pm$0.13 & 92.68$\pm$0.22 &  \underline{70.36$\pm$0.33} \\
    CAL & 91.97$\pm$0.32 & 90.93$\pm$0.31 & 90.75$\pm$0.30 & 90.74$\pm$0.22 & 85.36$\pm$0.16 & 61.73$\pm$0.42\\
    CORES & 91.23$\pm$0.11 & 89.66$\pm$0.32 & 89.91$\pm$0.45 & 89.79$\pm$0.50 & 83.60$\pm$0.53 & 61.15$\pm$0.73\\
    CORES$^{*}$ &  \underline{95.25$\pm$0.09} & 94.45$\pm$0.14 & 94.88$\pm$0.31 & 94.74$\pm$0.03 & 91.66$\pm$1.60 & 55.72$\pm$0.42\\ 
    CE & 87.77$\pm$0.38 & 85.02$\pm$0.65 & 86.46$\pm$1.79 & 85.16$\pm$0.61 & 77.69$\pm$1.55 & 55.50$\pm$0.66\\
    Negative-LS & 91.97$\pm$0.32 & 90.29$\pm$0.32 & 90.37$\pm$0.12 & 90.13$\pm$0.19 & 82.99$\pm$0.36 & 58.59$\pm$0.98\\
    SOP & $\mathbf{95.61\pm0.13}$ & 95.28$\pm$0.13 & 95.31$\pm$0.10 &  95.39$\pm$0.11 & $\mathbf{93.24\pm0.21}$ & 67.81$\pm$0.23\\
    \midrule
    \method\ (Best) & 95.03$\pm$0.14 & $\mathbf{95.48\pm0.17}$ & $\mathbf{95.48\pm0.21}$ & $\mathbf{95.67\pm0.10}$ &  \underline{92.97$\pm$0.33} & 67.73$\pm$0.67\\
    \method\ (Last)& 94.82$\pm$0.15 &  \underline{95.40$\pm$0.21}  &  \underline{95.37$\pm$0.33} &  \underline{95.53$\pm$0.09} & 92.48$\pm$0.18 & 67.44$\pm$0.22 \\
  \bottomrule
  \end{tabular}
\end{table*}

In this section, we evaluate the performance of \method\ by comparing it with representative noisy label learning methods on the datasets that are generated from real-world annotation scenarios. 

\paragraph{Dataset} 

\begin{table}[h]
  \caption{The statistics of CIFAR-N dataset}
  \label{tab:cifarn}
  \centering
  \resizebox{\linewidth}{!}{
  \begin{tabular}{c|c|c|c|c}
  \toprule
  Dataset & Subset & Train/Test size & Classes & Noise rate \\
  \midrule
  \multirow{5}*{\makecell[c]{CIFAR-10N}} & aggre & 50K/10K & 10 & 9.03\% \\
   & rand1 & 50K/10K & 10 & $\approx$18\% \\
   & rand2 & 50K/10K & 10 & $\approx$18\% \\
   & rand3 & 50K/10K & 10 & $\approx$18\% \\
   & worst & 50K/10K & 10 & 40.21\% \\
  \midrule
  CIFAR-100N & noisy & 50K/10K & 100 & 40.20\% \\
  \bottomrule
  \end{tabular}
  }
\end{table}
Following previous works~\cite{DBLP:conf/icml/LiuZQY22}, we use the CIFAR-N~\cite{DBLP:conf/iclr/WeiZ0L0022} dataset as the experimental benchmark, which has six subsets. CIFAR-N is created by relabeling CIFAR-10N/CIFAR-100N with Amazon Mechanical Turk labeling of the original CIFAR-10/CIFAR-100 dataset~\cite{krizhevsky2009learning}. The CIFAR-10N dataset contains five distinct noise rate options, including aggre, rand1, rand2, rand3, and worst. While the CIFAR-100N dataset only has one noise rate, namely noisy. Table~\ref{tab:cifarn} gives the statistics of all subsets in the CIFAR-N.

\paragraph{Baselines}
We choose the representative noisy label learning methods as the comparison methods:
\begin{itemize}
    \item Co-Teaching~\cite{DBLP:conf/nips/HanYYNXHTS18} trains two neural networks simultaneously and lets them teach each other.
    \item Peer-Loss~\cite{DBLP:conf/icml/LiuG20} introduces a new family of loss functions based on the standard empirical risk minimization framework.
    \item Divide-Mix~\cite{DBLP:conf/iclr/LiSH20} models the per-sample loss distribution to perform sample selection and improves MixMatch.
    \item ELR~\cite{DBLP:conf/nips/LiuNRF20} designs a regularization term to prevent memorization of the false labels. And ELR+~\cite{DBLP:conf/nips/LiuNRF20} introduces cosine annealing learning rate to enhance training based on ELR.
\item Positive-LS~\cite{DBLP:conf/icml/LukasikBMK20} introduces label smoothing and shows its effect is similar to label correction.
\item PES-semi~\cite{DBLP:conf/nips/BaiYHYLMNL21} separates the neural networks into different parts and progressively trains them.
\item CAL~\cite{DBLP:conf/cvpr/ZhuL021} proposes a new loss function based on covariance-assisted learning.
\item CORES~\cite{DBLP:conf/iclr/ChengZLGSL21} progressively sieves out corrupted examples from instance-dependent label noise. And CORES$^{*}$~\cite{DBLP:conf/iclr/ChengZLGSL21} applies consistency training.
\item  CE~\cite{DBLP:conf/iclr/WeiZ0L0022} constructs the CIFAR-N dataset and performs the base cross-entropy model. 
\item  Negative-LS~\cite{DBLP:conf/icml/WeiLL0SL22} proposes a negative weight to combine the hard and soft labels.
\item SOP~\cite{DBLP:conf/icml/LiuZQY22} proposes a principled approach for robust training of over-parameterized deep networks.
\end{itemize}

\paragraph{Implementation details}
We implement \method\ by PyTorch. For fairness, all experiments are performed on a single 1080Ti GPU. We use the ResNet34~\cite{DBLP:conf/cvpr/HeZRS16} as the backbone and optimize it by SGD with cosine scheduler. 
% Following previous work~\cite{DBLP:conf/iclr/WeiZ0L0022}, we choose the best test result on all epochs as the final prediction result. 
For each dataset, we run 5 times and report the mean and standard deviation. The learning rate, training epoch and weight decay are set to 0.02, 600, and 0.0005, respectively. And we perform a grid search to tune hyper-parameters: warm-up epoch $T_w \in \{10, 20, 30\}$, proto split epoch $T_p \in \{1, 2, 3\}$, lower threshold $\alpha \in \{0.99, 0.97, 0.95\}$, upper threshold $\beta \in \{0.94, 0.92, 0.90\}$. In semi-supervised learning, we apply random crop and flip to perform augmentation on the photos of CIFAR-N. The temperature for Mixup is set to $0.5$, and the balance factor $\lambda$ follows a linear decay value from the warm-up epoch to the training epoch.

\subsection{Experimental results}
Table~\ref{tab:res_main} shows the evaluation results on the CIFAR-N dataset. We take the results of other baselines from the leaderboard~\footnote{\url{http://competition.noisylabels.com/}} of CIFAR-N.
From the table, we observe: 1) \method\ achieves the best average performance on all CIFAR-N datasets. Notably, \method\ has surpassed the state-of-the-art (SOTA) results on the rand1, rand2, and rand3 datasets. This demonstrates the effectiveness of our approach for addressing the noisy label learning problem in real annotation scenarios. 
2) \method\ performs better on datasets with fewer classes, namely CIFAR-10N, as it can effectively construct prototype vectors for each category on these datasets.
Oppositely, the generated class-prototype vectors on datasets with more categories (CIFAR-100N) are difficult to distinguish. We will discuss this phenomenon in the further experiments (See the last two experiments). 
3) Sample selection-based methods, including Divide-Mix, PES-semi, and our proposed method \method\ all achieve relatively high performance, highlighting the importance of dividing the noisy dataset into confident dataset and unconfident dataset in noisy label learning.

\subsection{Ablation Study}

\begin{figure*}[h]
    \centering
    \includegraphics[width=0.95\textwidth]{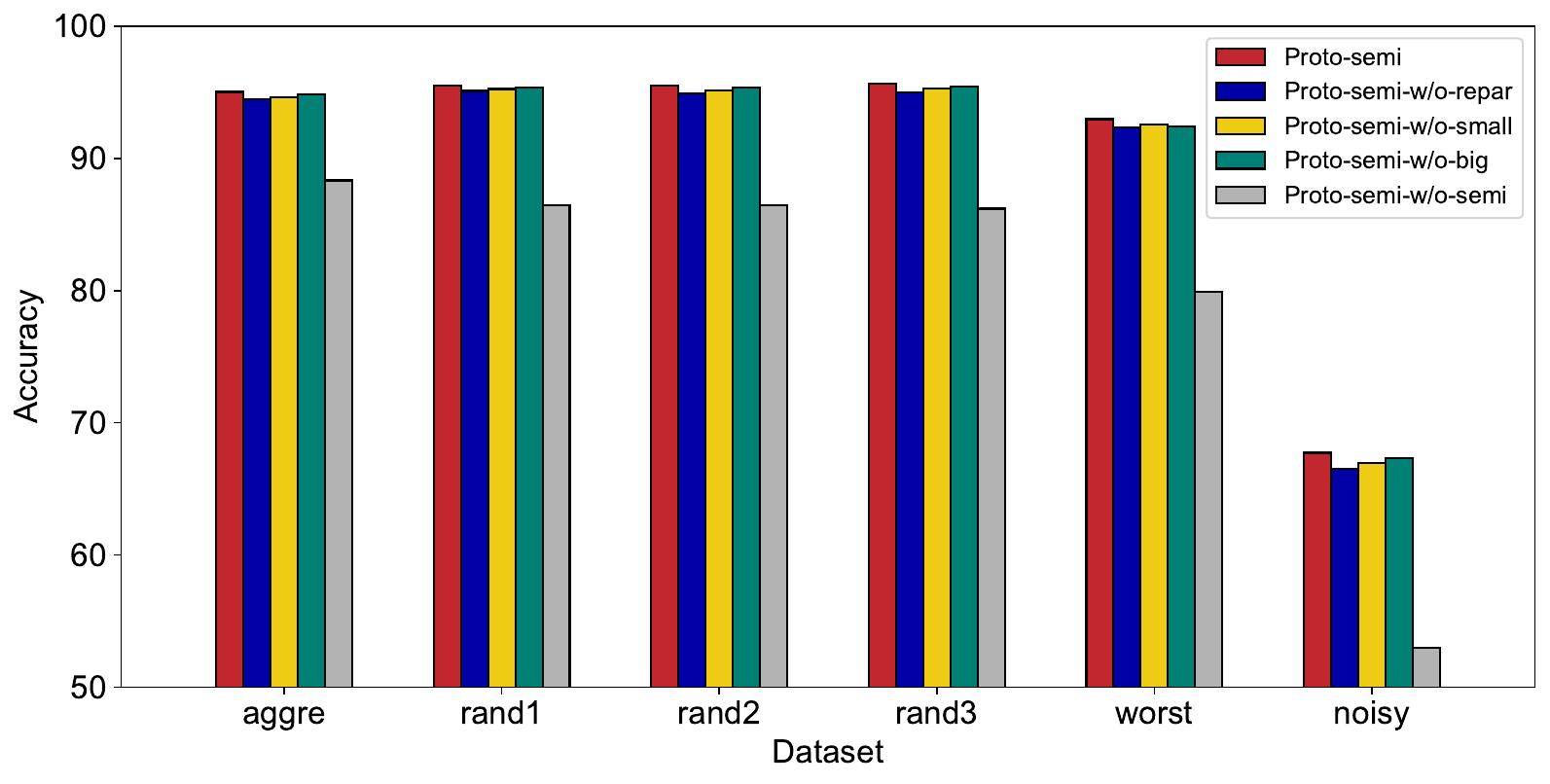}
    \caption{Ablation study on \method. From left to right: \method, \method-w/o-repar, \method-w/o-small, \method-w/o-big, \method-w/o-semi.}
    \label{fig:abl}
\end{figure*}
We further conduct ablation study to verify the importance of its main components in \method, including the repartitioning strategy in the sample selection, and the semi-supervised learning. 
As shown in Figure~\ref{fig:abl}, we implement four variants of \method\ by removing the corresponding component.

% The first variant, which we named \method-w/o-repar, removes the repartitioning strategy in the sample selection. From the comparison in Fig~\ref{fig:abl},  we can observe that there is a certain degree of decline in the results across all datasets. This indicates our proposed repartitioning strategy can generate better confident and unconfident datasets, which helps the model to achieve better prediction performance.

We name the first variant \textbf{\method-w/o-repar}, which removes the repartitioning strategy in the sample selection. As in Figure~\ref{fig:abl}, we observe declines in performance across all datasets. This suggests that our proposed repartitioning strategy can generate better confident and unconfident datasets, which ultimately helps \method\ achieve better classification performance. To further validate the impact of the re-partitioning strategy in detail, we conduct two ablation analyses and introduce two new variants: \textbf{\method-w/o-small} and \textbf{\method-w/o-big}. The former involves removing the construction of the small circle in the re-partitioning strategy, which means that we do not identify factual noise. On the other hand, the latter omits the construction of the large circle in the re-partitioning process, resulting in the failure to identify ambiguous noise.
The results in Figure~\ref{fig:abl} show that removing either circle construction leads to a decrease in the model's final performance compared to the complete method, indicating that both circle constructions are essential for the sample re-partitioning strategy. Moreover, on most datasets, removing the small circle construction has a greater impact on performance reduction than removing the large circle construction. This suggests that constructing small circles can effectively identify factual noises and correct their labels. Notably, on the CIFAR-10(worst) dataset, retaining ambiguous noise can improve the model's robustness and performance to some extent compared to removing the strategy that constructs small circles. The last variant, \textbf{\method-w/o-semi}, involves discarding the semi-supervised learning step and retaining only the warmup stage described in Algorithm~\ref{algorithm1}. 
Figure~\ref{fig:abl} shows a significant drop in performance across all datasets with this variant.
Furthermore, Table~\ref{tab:res_main} demonstrates that methods based on semi-supervised learning, such as PES-semi~\cite{DBLP:conf/nips/BaiYHYLMNL21} and Divide-Mix~\cite{DBLP:conf/iclr/LiSH20}, perform well on the CIFAR-N dataset.
These findings indicate that the training augmentation provided by semi-supervised learning is highly effective for the noisy label learning in real-world annotation scenarios.

\subsection{Prototype Embedding Visualization}\label{sec:proto_visual}

\begin{figure*}[h]
	\centering
	\subfloat[CIFAR-10N (rand2)]{\includegraphics[width=.98\columnwidth]{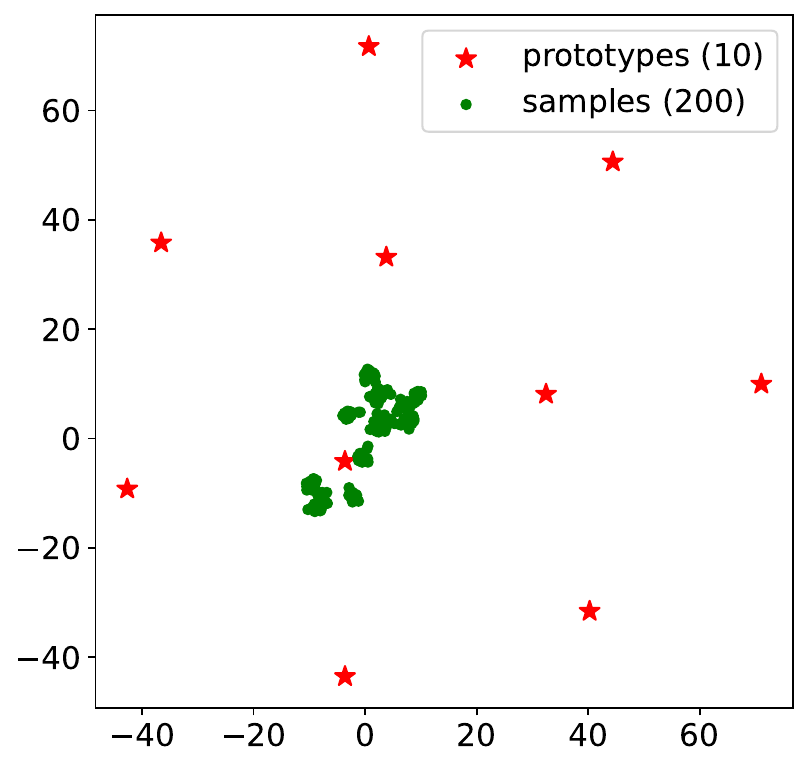}\label{fig:cifar10n_proto}}
 \quad
	\subfloat[CIFAR-100N (noisy)]{\includegraphics[width=.98\columnwidth]{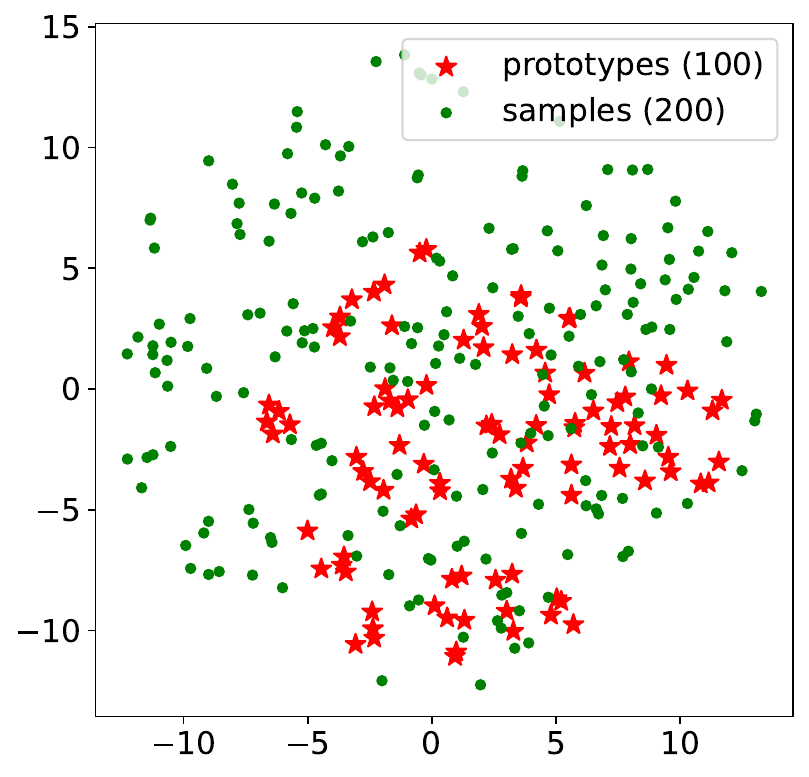}\label{fig:cifar100n_proto}}
	\caption{Visualization of prototype embedding on the CIFAR-N dataset.}
\label{fig:proto_visual}
\end{figure*}

\begin{table*}[h]
  \caption{The label correction results of small circle in \method\ ($T_w = 20, \alpha = 0.95$).}
  \label{tab:res_label}
  \centering
  \begin{tabular}{c|ccccc|c}
  \toprule
  & \multicolumn{5}{c}{CIFAR-10N}  &  CIFAR-100N  \\
   & aggre & rand1 & rand2 & rand3 & worst & noisy  \\
   \midrule
    % Number of samples  & & & & & & \\
    Size of unconfident dataset & 10534 & 10995 & 10611 & 5836 & 21946 & 25096\\ 
    Number of samples in small circle & 4015 & 5766 & 4258 & 1729 & 9472 & 1823\\
    Number of label corrections: & 3992 & 5703 & 4235 & 1713 & 9381 & 1708\\
    Right Number of label corrections & 3846 & 5416 & 4097 & 1666 & 8723 & 1325\\
    Correction accuracy & 96.34\% & 94.97\% & 96.74\% & 97.26\% & 92.99\% & 77.58\%\\
  \bottomrule
  \end{tabular}
\end{table*}

To validate the effectiveness of prototype vector construction and its performance on learning with noisy labels, 
we conduct dimensionality reduction and visualization experiments using t-SNE. 
We choose CIFAR-10N (rand2) and CIFAR-100N (noisy) as the experimental datasets.
Then a batch of 200 samples are randomly extracted from the unconfident dataset and perform visualization with all class prototype vectors. 
The visualization results are shown in Figure~\ref{fig:proto_visual}.
Figure~\ref{fig:cifar10n_proto} depicts the visualization result on the CIFAR-10N (rand2) dataset.
It can be observed that the 10 class prototype vectors were evenly distributed in the two-dimensional space,
and the 200 sample vectors were effectively clustered around a class prototype vector. This facilitated label correction or retention based on distance.
While Figure~\ref{fig:cifar100n_proto} shows the visualization result for the CIFAR-100N (noisy) dataset where the 100 class prototype vectors and 200 sample vectors were scattered without any discernible patterns in the two-dimensional space.
Some prototype vectors even overlap, making it challenging to reassign the samples based on these class prototype vectors.
Despite using high thresholds, we found it difficult to classify these samples in our CIFAR-100N experiment.

Overall, the visualization results in Figure~\ref{fig:proto_visual} indicate that our prototype-based repartitioning strategy can effectively learn class information in the dataset with few classes. By reassigning samples based on distances and correcting their labels, this strategy helps \method\ achieve good results in learning with label noise.

% \begin{figure}
%      \centering
%      \begin{subfigure}[b]{0.48\textwidth}
%         \centering
%         \includegraphics[height=6.5cm]{figs/cifar10_random_label2_1.pt_new.pdf}
%         \caption{CIFAR-10N}
%         \label{fig:cifar10n_proto}
%      \end{subfigure}
%      \hfill
%      \begin{subfigure}[b]{0.48\textwidth}
%         \centering
%         \includegraphics[height=6.5cm]{figs/cifar100_noisy_label_1.pt_new.pdf}
%         \caption{CIFAR-100N}
%          \label{fig:cifar100n_proto}
%      \end{subfigure}
%         \caption{Visualization of prototype embedding on the CIFAR-N dataset.}
%         \label{fig:proto_visual}
% \end{figure}

\subsection{Label Correction Analysis}\label{sec:label_correct}

In the last experiment, we analyze the effectiveness of the proposed prototype-based repartitioning strategy in correcting labels. 
We record several metrics, including the size of the unconfident dataset, the number of samples falling within a small circle (See Figure~\ref{fig:noise_type}), the number of samples corrected within the small circle, the number of samples correctly corrected, and the accuracy of correct corrections on all CIFAR-N datasets after the warm-up phase. The results are presented in Table~\ref{tab:res_label}, and the following discoveries can be drawn: 1) The strategy is highly effective in correcting labels. The accuracy of correction is higher than the model's final predictive performance on almost all datasets (See Table~\ref{tab:res_main}). Remarkably, the strategy achieves an impressive 97.26\% accuracy on the CIFAR-10N (rand3) dataset. 2) The majority of samples falling within the small circle require label correction due to factual noise. For instance, the CIFAR-10N (rand1) dataset and the CIFAR-100N (noisy) dataset have 5766 and 1823 samples within the small circle, respectively, with 5416 and 1708 samples needing correction. 3) The strategy performs better on datasets with fewer classes. This is shown by the lower number of unreliable samples, more samples within the small circle, and greater accuracy in label correction for all CIFAR-10N datasets compared to CIFAR-100N. These findings are consistent with the results presented in Figure~\ref{fig:proto_visual}.

\section{Conclusion}
In this paper, we present a novel sample selection-based approach, denoted as \method, for effectively identifying the type of noise and utilizing its semantics in real-world annotation scenarios. Specifically, \method\ divides the entire dataset into two subsets, namely confident and unconfident datasets, and subsequently leverages a distance-based technique based on prototype vectors to identify the types of annotation noise. Furthermore, the confident and unconfident datasets are adjusted based on the distances, and a semi-supervised learning strategy is incorporated to enhance the training process. Experimental results on the CIFAR-N dataset demonstrate the effectiveness of \method\ in addressing the problem of learning with noisy labels in real-world annotation scenarios. Avenue of future work will focus on extending our proposed method to more real-world annotation noisy datasets.

\bibliography{aaai24}

\end{document}